\documentclass[11pt]{article}

\usepackage{emnlp2022}

\usepackage{times}
\usepackage{latexsym}

\usepackage[T1]{fontenc}

\usepackage[utf8]{inputenc}

\usepackage{booktabs}
\usepackage{multirow}
\usepackage{float}
\usepackage[symbol]{footmisc}
\usepackage{tikz-network}
\usepackage{lipsum}
\usepackage{algorithm}
\usepackage[noend]{algpseudocode}
\usepackage{caption}
\usepackage{subfig}

\usepackage{graphicx}

\usepackage{amsmath,amsfonts,bm}

\def\eqref#1{equation~\ref{#1}}

\def\1{\bm{1}}

\DeclareMathAlphabet{\mathsfit}{\encodingdefault}{\sfdefault}{m}{sl}
\SetMathAlphabet{\mathsfit}{bold}{\encodingdefault}{\sfdefault}{bx}{n}

\renewcommand{\thefootnote}{\fnsymbol{footnote}}

\title{Deploying a Retrieval based Response Model for Task Oriented Dialogues}
\usepackage{graphicx}
\graphicspath{{../figures/}}

\author{Lahari Poddar $^*$ \And 
	Gyuri Szarvas $^*$ \And 
	Cheng Wang  \AND
	Jorge Balazs \\ \\ \\ \And
	 Pavel Danchenko \\ \\ Amazon \\
	\normalsize{ \texttt{\{poddarl, szarvasg, cwngam, jabalazs, danchenk, peernst\}@amazon.com} } \And 
	 Patrick Ernst \\ \\  \\
}

\begin{document}

\maketitle

\def\thefootnote{*}\footnotetext{These authors contributed equally}\def\thefootnote{\arabic{footnote}}

\begin{abstract}
Task-oriented dialogue systems in industry settings need to have high conversational capability, be easily adaptable to changing situations and conform to business constraints. This paper describes a 3-step procedure to develop a conversational model that satisfies these criteria and can efficiently scale to rank a large set of response candidates. First, we provide a simple algorithm to semi-automatically create a high-coverage template set from historic conversations without any annotation. Second, we propose a neural architecture that encodes the dialogue context and applicable business constraints as profile features for ranking the next turn. Third, we describe a two-stage learning strategy with self-supervised training, followed by supervised fine-tuning on limited data collected through a human-in-the-loop platform. Finally, we describe offline experiments and present results of deploying our model with human-in-the-loop to converse with live customers online.


\end{abstract}

\section{Introduction}\label{sec:intro}
\begin{figure}[!htb]
	\centering
	\subfloat[V0: Response Ranking with Poly-Encoder]{{\includegraphics[width=0.5\textwidth]{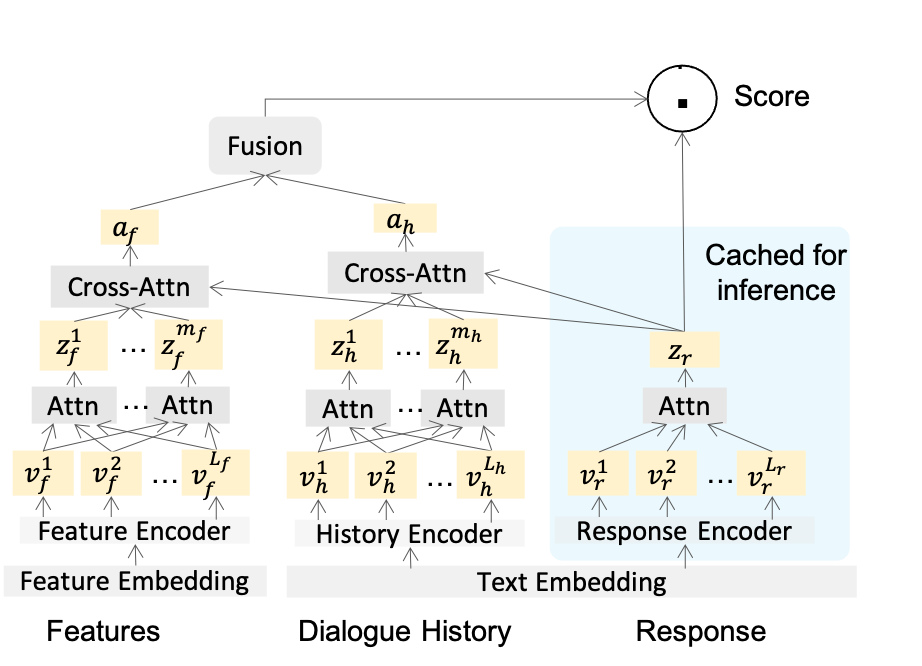} }}\\
	\subfloat[V1: Response Ranking with Shared Bert]{{\includegraphics[width=0.5\textwidth]{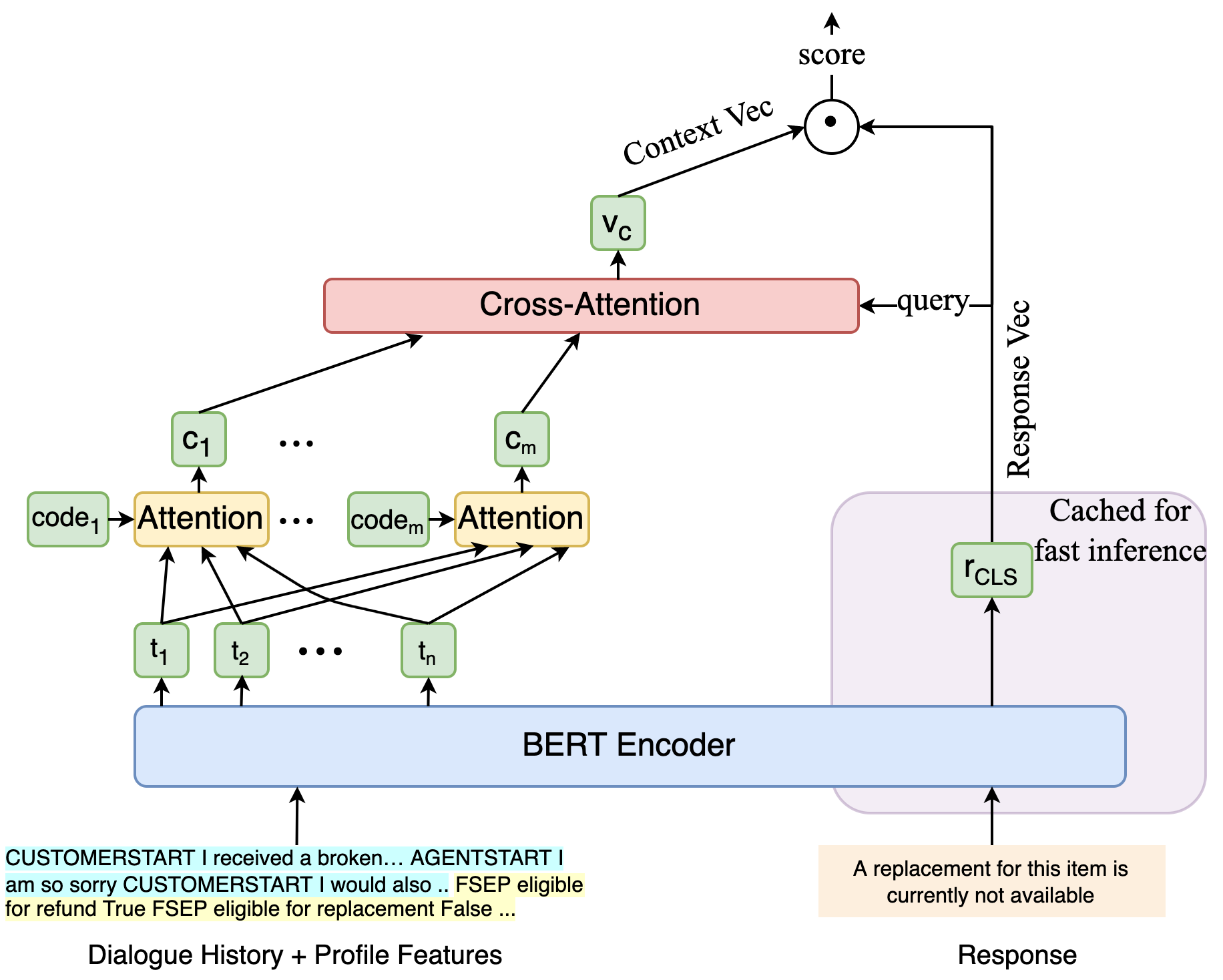} }\label{fig:bert_arch}}
	\caption{Production Ranking Models. The dialogue history, response and profile features are encoded with transformers (top) or using a shared Bert. Cross-attention layers learn the semantic correlation between history, features and candidate response. A score function computes and ranks candidate responses.}
	\label{fig:architecture}
\end{figure}

A Task Oriented Dialogue (TOD) system aims to accomplish specific tasks such as hotel reservation \cite{budzianowski-etal-2018-multiwoz}, flight booking, customer  support \cite{moore2021comprehensive} and so on.
An end-to-end TOD system directly takes a multi-turn dialogue context as input and predicts the next response with a single model~\citep{wen2016network}.
These can be developed using either retrieval-based approaches~\citep{ijcai2021-0627,chen2017survey} where the model ranks a response from a pre-constructed response pool; or generative approaches where a response is sequentially generated with encoder-decoder architectures ~\citep{serban2017hierarchical,sordoni2015neural}.
Although generative models are widely studied in literature for dialogue systems \cite{hosseini2020stod,yang2021ubar} as they are capable to generate free text, it is nearly impossible to provide guarantees on the style, quality and privacy risks for their real-world applications.

In this work, we focus on the development and deployment of a retrieval-based conversational system for an online retail store, in the customer service domain.

Our main contributions are:


\begin{enumerate}
\item We design a simple yet effective algorithm for generating a large, representative response pool from un-annotated dialogues and show that it can achieve high coverage for handling natural language conversations.
\item We present an approach which combines self-supervised training (from human-human conversations) and supervised fine-tuning (from human-in-the-loop interactions) for learning dialogue models in real industry settings.
\item We enhance state-of-the-art Poly-Encoders architecture for retrieval based dialogue system, incorporating multi-modal information from dialogue text, and non-textual features associated with the order and the customer.
\item We present a breakdown of development and deployment stages of the conversational system from offline evaluation --> controlled human-in-the-loop setting --> fully online on live traffic with real customer contacts.
\end{enumerate}

\section{Related Work}
Retrieval-based dialogue systems \cite{ijcai2021-0627} involve single- and multi-turn response
matching~\citep{chen2017survey, Lu2019GoalOrientedEC, Henderson2019TrainingNR,Gu2020SpeakerAwareBF,Whang2020AnED, poddar-etal-2022-dialaug,Xu2021LearningAE,vig2019comparison}. The selection of an appropriate response is usually based on computing and ranking the similarity between context and response.
Two popular model architectures for such similarity computation between inputs, is Cross-encoders \cite{DBLP:journals/corr/abs-1901-08149}, which perform full self-attention over a given input and label candidate; and Bi-encoders \cite{dinan2018wizard}, which encode the input and candidate separately and combine them at the end for a final representation. 
Bi-encoders have the ability to cache the encoded candidates, and
reuse their representations for fast inference. Cross-encoders, on the other hand, often achieve higher accuracy but are prohibitively slow at test time. A recent method, Poly-encoders \cite{humeau2019poly}, combines the strengths from the two architectures, and allows for caching response representations while implementing an attention mechanism between context and response for improved performance.
Transformer-based architectures~\citep{vaswani2017attention, devlin2018bert} are widely used to encode information in TOD systems. For instance, TOD-BERT~\citep{wu2020tod} incorporates user and system tokens into the masked language
modeling task and uses a contrastive objective function to simulate the response selection task. In this work, we also adapt the Transformer architectures and enhance Poly-Encoders to encode conversational history, response and profile features.

\section{Response Pool Creation}\label{sec:tempgen}
\begin{figure}[tbp]
	\centering
	\includegraphics[width=\linewidth]{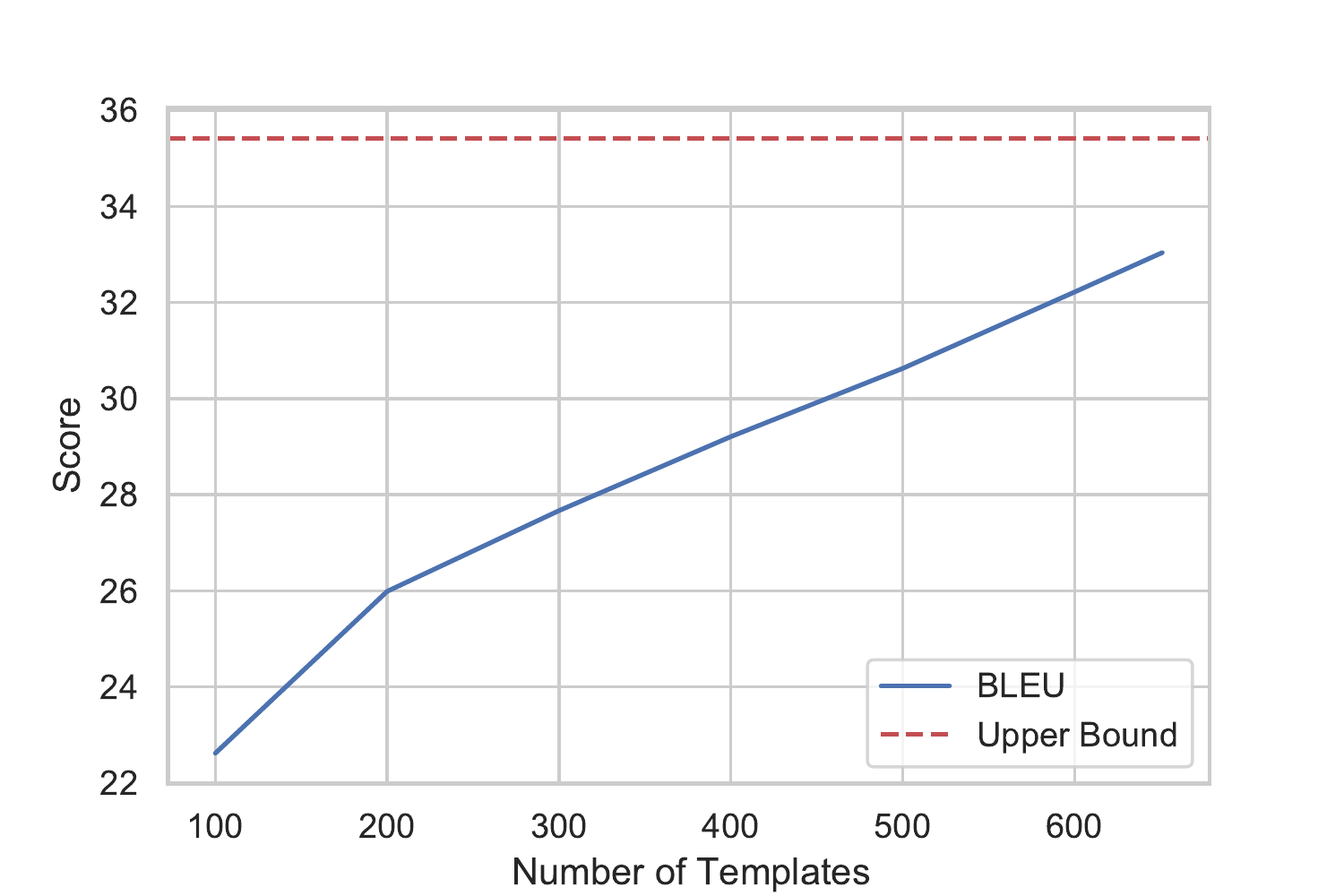}
	\caption{Template coverage on general conversations for Return Refund intent. Upper bound is established by adding templates to the pool based on human expert suggestions through several months of active use. }
	\label{fig:tempcov}
\end{figure}

We semi-automatically extract a broad template pool from a large number of anonymized human dialogues.
We first select the template texts from human responses in actual dialogues. This ensures that the bot language conforms to the desired style.

Our primary selection criteria for response candidates are frequency and novelty. We iteratively select sentences that are (1) most frequently used in human dialogues, and (2) contain information different from already selected responses (detailed algorithm in Appendix \ref{sec:appa}). This directly maximizes the dialogue model's coverage, as measured by the fraction of contexts for which the model has a suitable response in the pool.
An alternative approach would have been clustering frequent sentences and selecting a representative for each cluster~\cite{Hong2020EndtoEndTD} as templates. We instead opted for the deterministic procedure which is more intuitive for ingesting prior linguistic knowledge and provides interpretability.

\noindent \textbf{Quantitative Evaluation of Coverage}: Figure \ref{fig:tempcov} shows the BLEU score by aligning the best matching templates to unconstrained human-human dialogues. As can be seen, with the growing size of the template pool, the BLEU score approaches this upper bound, proving that the proposed approach can achieve strong conversational capacity. 
Similar to ~\citep{swanson-etal-2019-building}, we see that a set of $500-1$k sentences can achieve good coverage of domain-specific conversations.

\noindent \textbf{Template Decorations}: We enhance the templates through attaching metadata, like calls to external APIs and constraints on profile features.
For example, a template `I have issued a refund to your credit card', will have an action that triggers an API call for issuing the refund.
Through \texttt{profile feature constraints} we enforce consistency requirements on the dialogue, for example, filtering out the above template if an order is not eligible for refund. This establishes guarantees that the bot is always consistent with business policies.

\section{Model Architecture}
We represent a dialogue $D_i=\{a_1, u_1,  \ldots, a_n \}$ as a set of user ($u_i$) and agent ($a_i$) turns.
While conversing with a user, agents look up information related to the particular order and item to determine applicable business policies and constraints. This may include item category, its delivery status, whether it was already refunded, among others; we encode these as categorical features.

We create multiple input-output tuples by splitting a complete dialogue transcript at each agent turn (e.g.~at turn $k$). The model learns to predict the next agent response ($a_k$) given the dialogue history so far and the features that encode item level, customer level information and applicable policies.
We flatten the history into a single sequence by concatenating all agent and user turns ($x_h$). We introduce two marker tokens [AGENTSTART] and [USERSTART] to mark the beginnings of an agent and user turn respectively. The features are also represented as a sequence, where each feature is encoded as a $key\_value$ pair ($x_f$).

Figure \ref{fig:architecture} presents the overall architecture of our proposed ranking model extended from Poly-Encoder
\cite{humeau2019poly}. We use separate transformers \cite{vaswani2017attention}  as encoder blocks ($\mathcal{M}$) for all inputs and encode them as:
 \begin{align}
\mathbf{v}_h=\mathcal{M}_{h}(x_h), \mathbf{v}_h  \in \mathbb{R}^{L_h \times d} \\
\mathbf{v}_f=\mathcal{M}_{f}(x_f), \mathbf{v}_f  \in \mathbb{R}^{L_f \times d} \\
\mathbf{v}_r=\mathcal{M}_{r}(x_r), \mathbf{v}_r  \in \mathbb{R}^{L_r \times d}
\end{align}

\noindent where $d$ is the dimension of the output vector, and $L_h, L_f, L_r$ are maximum sequence lengths of history, features and responses respectively, and $x_r$ is the target response.

Over the sequences we apply self-attentions to obtain latent representations. We represent the response using a single vector $\mathbf{z}_r \in \mathbb{R}^d$.
For the multi-turn dialogue context and feature sequence we learn $m_h$ and $m_f$ representations, respectively, i.e. $\mathbf{z}_h \in \mathbb{R}^{m_h \times d}$ and $\mathbf{z}_f \in \mathbb{R}^{m_f \times d}$. We use $300$ history representations ($m_h$) and  $50$ profile representations ($m_f$) in our experiments.

To learn history-response and feature-response correlations we apply cross-attention layers.
 \begin{align}
\mathbf{a}_h = Att_{cross}(K=\mathbf{z}_h, V=\mathbf{z}_h, Q=\mathbf{z}_r) \\
\mathbf{a}_f = Att_{cross}(K=\mathbf{z}_f, V=\mathbf{z}_f, Q=\mathbf{z}_r)
\end{align}
where $K, V, Q$ present the key, value, query respectively, $\mathbf{a}_h \in \mathbb{R}^{d}$ and  $\mathbf{a}_f \in \mathbb{R}^{d}$ are the final history and profile representations.

We then merge the two modalities of information from history and profile features through a 2-layer MLP  to represent the complete dialogue context:
\begin{equation}
\mathbf{a}_{hf} = \mathcal{F}([\mathbf{a}_h, \mathbf{a}_f])
\end{equation}
A score function is used to rank the candidate responses given a (history, profile) pair through computing similarity using dot-product.
 \begin{equation}
\mathbf{s}= f_{score}( \mathbf{a}_{hf},  \mathbf{z}_r)
\end{equation}

We train the model in end-to-end manner using a binary cross-entropy loss.

\label{sec:model}

\section{Model Development}
In order to protect customer experience and trust, we do not simply train a model on human-human conversation data and deploy it to live traffic directly. To utilize the expertise of our customer service agents, we introduce a subsequent stage that not only acts as an intermediate test-bed, but also provides a fly-wheel to annotate data. Our model training consists of the following two stages.

\subsection{Self-supervised Training}\label{subsec:pretrain}
A large volume of anonymized human-human conversations is used for learning an initial dialogue model via Self-Supervised Training (SST). The goal is to rank the correct next utterance higher compared to other randomly sampled utterances given the dialogue history and associated profile features. Note that this model is independent of the response pool discussed in Section \ref{sec:tempgen}.

\subsection{Supervised Fine-Tuning}
\label{aab}
We use the model obtained from the previous stage to collect supervision data within a human-in-the-loop environment.
In this setting, whenever a customer starts a contact, the utterance, along with profile features, is passed on to the SST model. The entire response pool is ranked by the model and top $k$ responses are shown to human agents. They have three options for responding to the customer- a) accept the suggestion, b) pick a different template from the pool, c) indicate a failure of the pool (no response in the pool can be used to progress the conversation) or the constraints (e.g.~a refund should be offered but it is not available).


We utilize the data collected through this human-in-the-loop setup for further Supervised Fine-Tuning (SFT) of the model. The key difference compared to the previous stage is that in this setup
both the positive and negative responses come from the response pool.
We create a set of $N$ candidates with the one that the human expert accepted or searched as positive. The negative candidates are sampled randomly from the template pool. Whenever the response used by the expert was obtained through search, we leverage the model-suggested responses as hard negatives.

The primary goal of this human-in-the-loop stage is to collect the best possible data for supervised training. Instead of the straightforward approach of suggesting the top-scored template to human experts, we found that a sampling strategy among high scoring templates can boost impressions for less frequent templates. This helps improving the utility of the collected data. \footnote{For space constraints, the implementation details and experimental results are found in Appendix \ref{sec:appc}}

\section{Evaluation}
We conduct offline and online experiments on internal conversational datasets of an e-commerce customer service from the following two intents,

\noindent 1. \textbf{Start-Return (SR)}: where a customer wants to initiate a return of an item.

\noindent 2. \textbf{Return-Refund Status (RRS):} all post-return cases where customer may enquire about the status of a return or refund already issued / currently under processing.


\subsection{Experimental Setup}

\begin{table}[tbp]
	\resizebox{\linewidth}{!}{%
		\begin{tabular}{lllll}
			\toprule
			 & \multicolumn{2}{c}{Start Return} & \multicolumn{2}{c}{Return Refund} \\
             \cmidrule(lr){2-3} \cmidrule(lr){4-5}
			& Unlabeled & Labeled & Unlabeled & Labeled \\
             \midrule
			\# Dialogues & 824K & 30K & 918K & 21.6K \\
			Avg. \# Turns & 18.9  & 13.5 & 30.6 & 8.7  \\
            \bottomrule
		\end{tabular}%
	}
    \caption{Overview of datasets.}
	\label{corpus-stats}
\end{table}

\textbf{Datasets}. The dataset statistics are summarized in Table \ref{corpus-stats}. We tokenize and sentence split dialogue turns using NLTK toolkit \cite{loper2002nltk}. We split each dataset to train/dev/test sets with ratio 90:5:5 and use the most frequent $30$K, $10$K tokens as dialogue encoder vocabulary for Start Return and Return Refund intent, respectively.

\noindent\textbf{Model Training}. We train and fine-tune the models on the unlabeled and labeled intent datasets, respectively. We use a learning rate of $0.00015$ and train for $30$ epochs with early stopping.

\noindent\textbf{Metrics}. For offline evaluation, we use the standard ranking metrics Recall@$k$, and MRR (Mean Reciprocal Rank), and a metric for offline manual evaluation for top scored template, namely,

\noindent 1. \emph{Template Precision (TP)}: 
For 200 samples drawn randomly from test set, we use the model to rank templates in the pool. We manually evaluate the acceptability of the top-ranked template by the model and report an averaged precision.

For online evaluation we introduce:

\noindent 2. \emph{Turn-level Acceptance Rate (TAR@k)}: $\frac{N_{accept}}{N_{total}}$, $N_{accept}$ is the number of turns accepted by human expert out of the number of total turns $N_{total}$. TAR is an online correspondent of the $Recall@k$ metric. A higher value of TAR indicates model's capability of handling a conversation well, through ranking of the template pool.

\noindent 3. \emph{Task Completion (TC)}: The percentage of contacts that agents were able to resolve - either by accepting model suggestion or searching the pool. TC measures the quality and capacity of the pool and sets an upper bound for the bot's success rate.

\noindent 4. \emph{Automated Task Completion (ATC)}: Success rate of the deployed bot; i.e.~the percentage of contacts where the system is able to resolve the customer issue, such that the same customer doesn't repeat the contact within next 24 hours.

\subsection{Self-Supervised Training Results}

We first report offline and online results of models trained using human-human dialogues in self-supervised manner.
We deploy the trained model in online human-in-the-loop setup (described in Section \ref{aab}) and measure TAR@k and TC. We share our key learnings in this section.

\begin{table}[tbp]
	\resizebox{\linewidth}{!}{%
		\begin{tabular}{ccccccc}
			\toprule
			\multirow{2}{*}{Intent} & \multicolumn{3}{c}{Offline Metrics} & \multicolumn{3}{c}{Online Metrics} \\
			\cmidrule(lr){2-4} \cmidrule(lr){5-7}
			& Rec@1/29 &MRR & TP & TAR@4 & TAR@1 & TC \\
			\midrule
			SR  & 76\% & 86\% & 76\% & -- & 71\% & 52\% \\
			RRS & 71\% & 81\% & 71\% & 50\%& 17\%$^{\dagger}$ & 39\% \\
			\bottomrule
		\end{tabular}%
	}
	\caption{Offline and Online results for initial dialogue model trained with self-supervision. \footnotesize{$^{\dagger}$ RRS was launched in top-4 suggestion mode, while the better performing SR intent was launched in top-1 suggestion mode.
		}
	}
	\label{base-eval}
\end{table}

\begin{figure*}[htbp]
	\centering
	\subfloat[\label{aab_only_training}SFT]{{\includegraphics[width=0.25\textwidth]{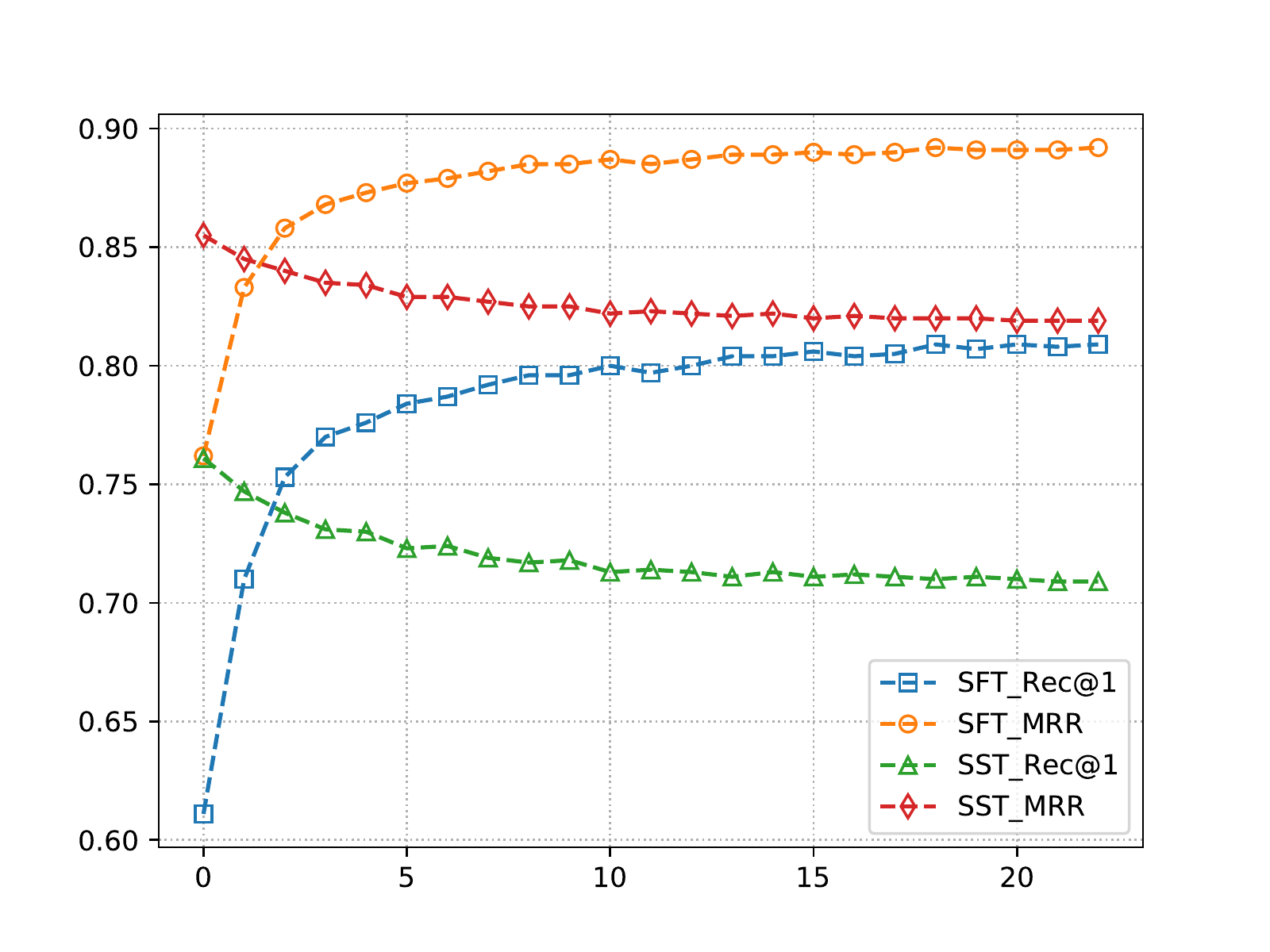} }}
	\subfloat[\label{balanced_training}SFT+SST]{{\includegraphics[width=0.25\textwidth]{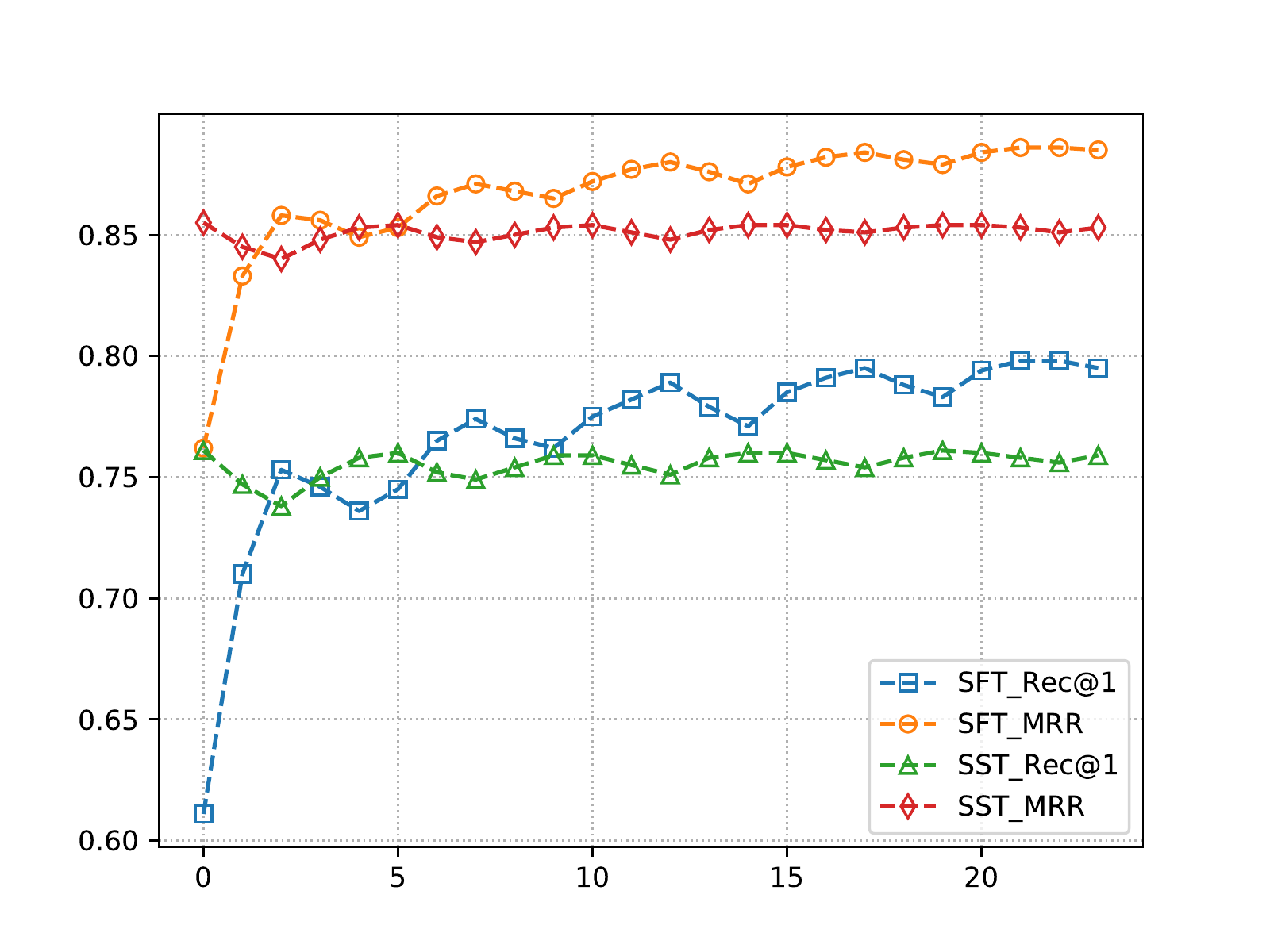} }}
	\subfloat[\label{subfig:rrs_aab_only_training}SFT]{{\includegraphics[width=0.2375\textwidth]{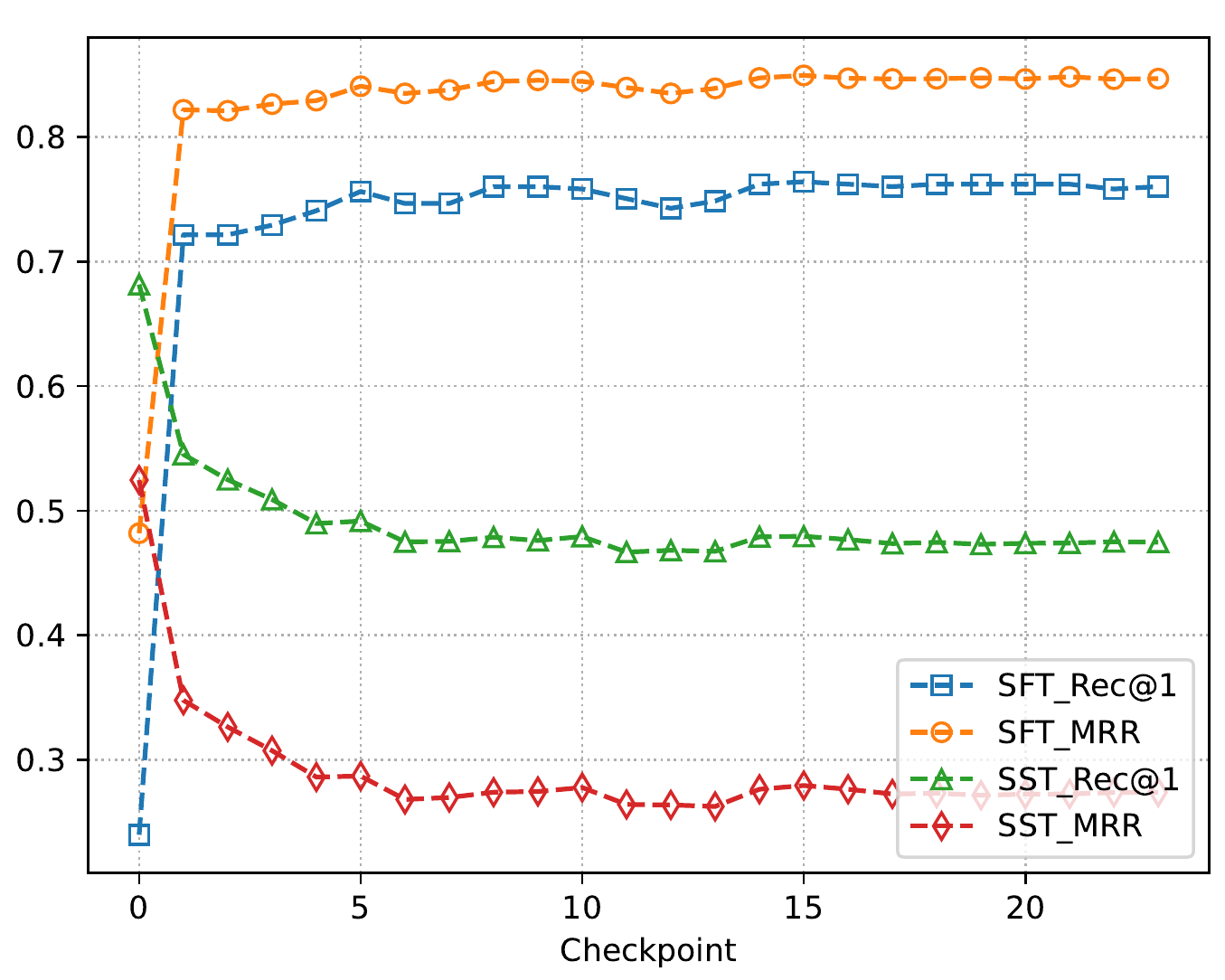} }}
	\subfloat[\label{subfig:rrs_balanced_training}SFT+SST]{{\includegraphics[width=0.2375\textwidth]{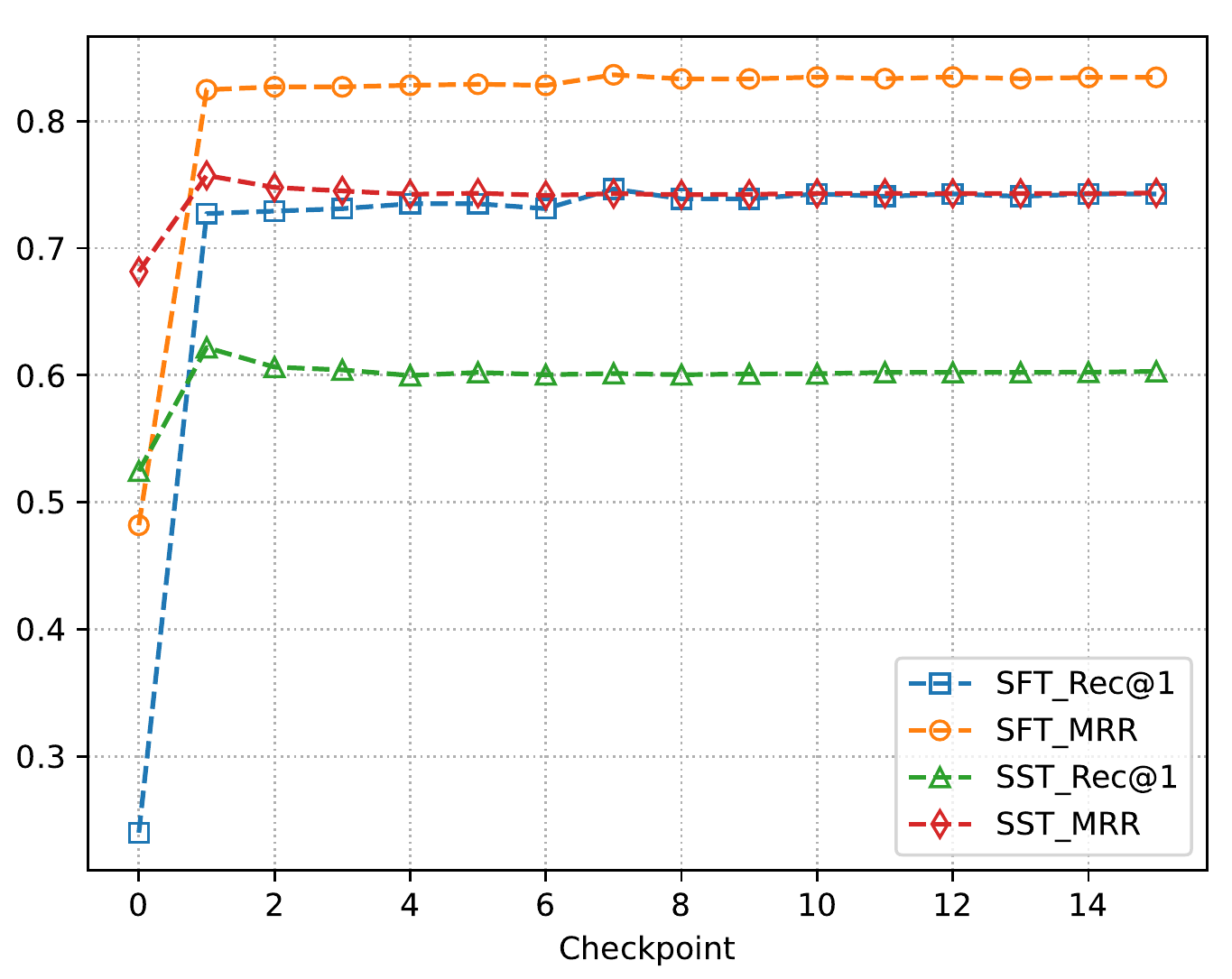} }}
	\caption{Evaluation after fine-tuning with labels collected from human-in-the-loop platform on the SR dataset ((a) and (b)) and the RR dataset ((c) and (d)). X-axis: checkpoints, Y-axis: performance.}
	\label{aab_results}
\end{figure*}

\noindent \textbf{Performance varies depending on domain complexity: } From the results in Table \ref{base-eval}, we first observe the significant gap in online metrics between the two intents, which shows that the performance of dialogue systems in real-world conversations are highly dependent on the complexities of the domain. This primarily reflects in the lower task completion rates (TC) of RRS, where dialogues often become open-ended when discussing issues with a previous return, compared to the more procedural dialogues about starting a return.

\noindent \textbf{Human choices are often arbitrary among close alternatives: }
In the RRS intent we conducted the online experiment by displaying top-$4$ templates to the agents instead of a single one. This decreases TAR@1 by \textasciitilde{15\%}.
Using more suggestions generally improves the agent's productivity due to limited search. However, we observed that human choices among similar templates are often arbitrary, leading to performance drops.

\noindent \textbf{Features are indisposable:} For RRS the TAR@4 was also quite low, especially given that SR had TAR@1 above 70\%. The main reason was the lack of crucial features, like granular tracking information from the carrier companies about the return package. This limited its ability to accurately condition on external factors compared to human experts.
This underlines the saliency of features (or external knowledge) in a practical TOD setting.


\noindent \textbf{Data-driven templates enable transfer learning}: From the offline results we note that the manually annotated TP metric closely resembles Recall@1 for both intents. This implies that the model is able to learn from  human-human conversations and apply it for ranking the restricted template set. Having a large template pool that follows a similar data distribution as the original agent responses helps in achieving this smooth transition.

\noindent \textbf{Large template pool is effective for handling conversations at scale}: The contact-level metric (TC) shows that with the generated template pool $52.1\%$ contacts could be fully resolved for the SR intent. This demonstrates the potential of using a large representative set of agent responses for tackling in-domain task oriented conversations.

\subsection{Results After Supervised Fine-Tuning}\label{subsec:finetuning-eval}

We explore two fine-tuning strategies with the limited data collected from human-in-the-loop stage.

\noindent \textbf{Catastrophic Forgetting with training only on restricted language:}  Figure \ref{aab_only_training} shows that fine-tuning with only the limited supervised data leads to better performance on the supervised test set (SFT) but increasingly worse performance on the general conversation test set (SST) as training progresses. This implies that as the model is being trained on this restricted data distribution, it is `forgetting' previously learned knowledge through self-supervision.
To mitigate this, we adopt a simple replay mechanism \cite{rolnick2019experience}. We augment the fine-tuning dataset by mixing in equal number of training instances from the self-supervised dataset. As seen from Figure \ref{balanced_training}, training with the balanced dataset leads to consistently better results on both datasets. 
This proves the ability of the model to learn from the limited supervised dataset without overriding previous knowledge.
Similar results were observed for RR intent Figure \ref{balanced_training} ((c)-(d))

\begin{table}[tp]
	\centering
	\small
	\begin{tabular}{ccccc}
		\toprule
		\multirow{2}{*}{Intent} & \multicolumn{2}{c}{Offline Metrics} & \multicolumn{2}{c}{Online Metrics} \\
		\cmidrule(lr){2-3} \cmidrule(lr){4-5}
		& Rec@1/29 &MRR & TAR@1 & TC \\
		\midrule
		SR  & 80.9\% & 89.1\% &  84.9\% & 56.6\% \\
		RRS & 76.4\% & 84.9\%  & 46.6\% & 46.3\% \\
		\bottomrule
	\end{tabular}
	\caption{Offline and Online results post finetuning}
	\label{finetuning-eval}
\end{table}

\noindent \textbf{Supervision from human-in-the-loop significantly boosts performance:} Table \ref{finetuning-eval} shows the performance after supervised fine-tuning.
In this experiment, the offline test set contains template responses from the  human-in-the-loop setup, in contrast to the general conversation responses considered in offline evaluation in Table \ref{base-eval}.
Offline metrics for both intents are generally higher compared to the general test set (Table \ref{base-eval}). This is expected, since by restricting to specific template set the language variation in agent turns is greatly reduced, making the ranking task easier.
More importantly, we observe the increase in the online metric TAR post fine-tuning, demonstrating the effectiveness of the two-stage training strategy.

\noindent\textbf{Human-in-the-loop setup also augments the template pool}: The increase in TC is independent of training strategies and fueled by enhancements to the template pool using suggestions from the human experts engaged in the supervised data collection process.
It is noteworthy that the initial (automatically collected) template pool attained a high 92\% and 84\% of the TC compared to these refinements. This demonstrates the efficacy of the proposed template creation method (Section \ref{sec:tempgen}).

\subsection{Deployment}
\begin{figure}[tbp]
	\centering
	\includegraphics[width=\linewidth]{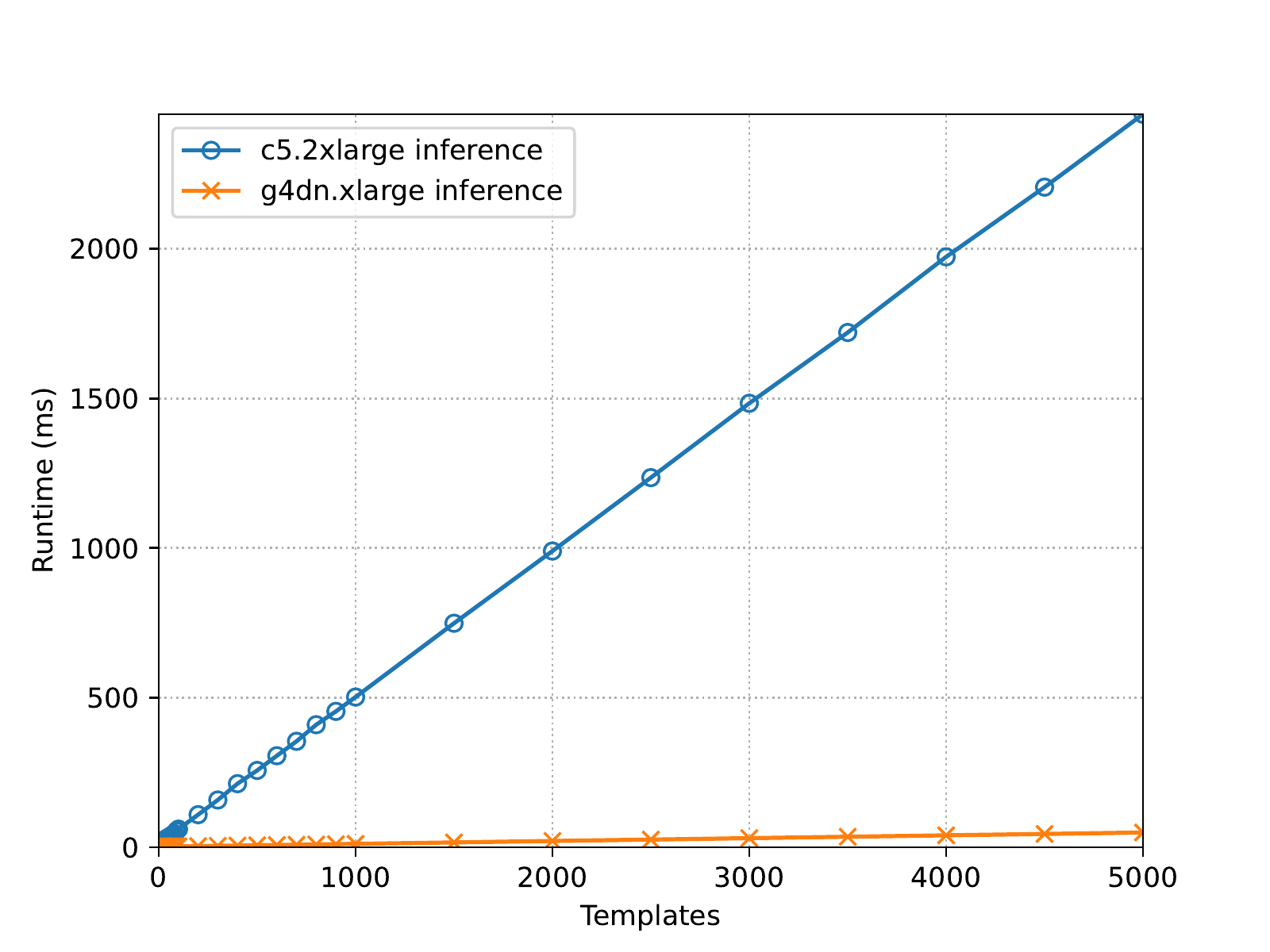}
	\caption{Inference speed with growing template pool}
	\label{fig:runtime}
\end{figure}

A key advantage of the proposed model architecture is its inference efficiency. 
The textual representation of templates can be encoded once at initialization and cached for future calls.
For each template only the cross-attention layers and final scoring needs to be re-computed.
This lets the inference latency scale linearly with template set size (Figure \ref{fig:runtime}). On {\tt c5.2xlarge} CPU instances with $1k$ templates the latency is below $0.5$ sec, which is sufficient for real time conversation with  users;
on a small GPU instance {\tt g4dn.xlarge} up to $5k$ templates can be scored within $50$ ms.

While the results after supervised fine tuning for SR reach sufficiently high quality for deployment of the chatbot,
additional improvements are needed for RRS, especially in providing the essential package tracking information through features. The Start Return chatbot achieved 48.3\% Automated Task Completion (ATC) after deployment.


\section{Developing with Pre-trained Language Models}
We continue developing and improving our models after deployment. With the initial model launched, we explore plugging in pre-trained language models like BERT as our text encoder (Figure \ref{fig:bert_arch}). We adopt a single shared encoder to simplify our architecture and limit the memory footprint of training and hosting a large model. Additionally, we convert the item features to have more semantic names (e.g. `eligible for refund') and append them to dialogue history for generating the complete context. This allows the self-attention mechanism of the transformer module to capture multi-modal interactions between features and dialogue turns that are grounded on them. 
We use batch negatives during training and learn the next response prediction task using categorical cross-entropy loss.

Table \ref{offline-bert} shows the offline evaluation results (analogous to Table \ref{base-eval}) for the two intents. Using PLMs clearly help in improving performance at a much higher data efficiency - with only 20\% of training data the BERT initialized model outperforms our previous model, which did not use any pre-training but was trained on 100\% available in-domain dataset. 
We further fine-tune the models using the supervised data collected through the human-in-the-loop setup described in Section \ref{aab}.

Next, we deployed the improved BERT model for SR intent to live customer traffic. Similar to the offline results, we observed a significant improvement in ATC to $55.3\%$ for the BERT model compared to $48.3\%$ for our previous model that did not use any pre-trained language model.
In future we plan to deploy for the RRS intent as well and compare both performance and efficiency.

\begin{table}[]
	\resizebox{\linewidth}{!}{%
		\begin{tabular}{cccccc}
			\hline
			\multirow{2}{*}{Data Size} & \multirow{2}{*}{Model} & \multicolumn{2}{c}{SR} & \multicolumn{2}{c}{RRS} \\ \cline{3-6}
			&  & Rec@1 & MRR  & Rec@1 & MRR \\ \hline
			100\% & no PLM & 76.1\% & 85.5\% & 71.2\% & 81.0\%  \\ \hline
			20\% & bert-base & 83.2\% & 89.5\%  & 76.3\% & 85.1\% \\ \hline
			50\% & bert-base & \textbf{88.2\%} & \textbf{92.9\%}  & \textbf{81.4\%} & \textbf{88.7\%} \\ \hline		
		\end{tabular}%
	}
	\caption{Offline performance comparison for pre-trained language model as text encoder}
	\label{offline-bert}
\end{table}

\section{Conclusion}
We presented a neural, retrieval-based dialogue model that ranks responses from a large, data-driven template pool.
Pre-defined responses make it possible to enforce requirements for consistency to business policies and the proposed template mining method provides good conversational capacity.
The model is accurate and efficient in terms of inference speed to handle conversations in real time.
A human-in-the-loop setup lets us effectively collect a small-sized labeled dataset to improve the quality for online deployments.

Offline and online results demonstrate that this is a viable approach for developing TOD systems for practical usecases. While RRS showed good improvements with our training protocol, it needs further work to be deployed. Performance on the SR intent permitted the deployment of the model; its live success rate almost reaches the 56\% upper bound that humans achieve in the controlled setting. Anecdotal evidence\footnote{We include few positive user feedback in Appendix \ref{sec:appd}.} from customer feedback shows that successful dialogues by the model provide good conversational experience.

\newpage
\section*{Ethical considerations}
\textbf{Development and experiments.} We used anonymized text dialogue snippets to train the models. The system predicts template responses, hence the model described in this work has no way to reveal customer information. This is actually a key theoretical advantage to generative models. We do not release the datasets used in the experiments.

\textbf{Failure modes.} Regarding risks related to system errors, incorrect predictions of the models described in this work may result in a confusing dialogue experience for customers. However, the practical risk related to such confusion is limited, because the chatbot operates in a semi-automation setting where it naturally predicts and transfers the contact to a human expert upon a drifting dialogue history. Moreover, customers also have an option to talk to a human associate upon request, if they consider the system doesn't work as expected.

\bibliographystyle{acl_natbib}
\bibliography{references}

\clearpage
\newpage

\appendix
\onecolumn
\section{Response Pool Creation} \label{sec:appa}
\begin{table*}[htpb]
\resizebox{\linewidth}{!}{%
\scriptsize
\begin{tabular}{llllllllll} \toprule
a-to-z & book & confirmation & dropoff & hold & name & price & reimbursement & security & transaction \\ 
accept & box & contact & e-mail & id & notification & print & reorder & sell & transfer \\ 
access & business & correspondence & elaborate & ignore & number & priority & repeat & seller & transit \\ 
account & call & cost & email & inconvenience & option & problem & replace & sender & understand \\ 
action & cancel & create & error & inform & order & proceed & replacement & service & understanding \\ 
address & card & credit & escalate & information & pack & process & representative & ship & update \\ 
allow & carrier & cvc & exception & initiate & package & product & request & shipment & ups \\ 
alternative & center & damage & exchange & inventory & packaging & promo & require & shipping & url \\ 
amount & certificate & delay & expedite & investigate & party & promotion & research & solution & use \\ 
apologize & charge & deliver & experience & investigation & patience & provide & resolution & specialist & verify \\ 
apology & check & delivery & expire & issue & pay & purchase & resolve & status & visa \\ 
apply & checking & department & extend & item & payment & qr & responsibility & stay & wait \\ 
arrange & claim & detail & fee & label & perfect & quantity & restock & stock & waive \\ 
arrive & click & device & feedback & leadership & phone & re-order & restocking & store & warehouse \\ 
assistance & code & digit & find & link & photo & reason & resubmit & subscription & warranty \\ 
associate & come & disarm & follow & locker & pick & receipt & retrocharge & suggest & website \\ 
authorization & compensation & discount & fulfil & mail & pickup & receive & return & supervisor & window \\ 
availability & complaint & display & fulfillment & mailing & picture & refer & returnable & support &  \\ 
balance & complete & dispose & fund & manufacturer & place & reflect & review & team &  \\ 
bank & concern & disregard & gift & member & policy & refund & safety & time &  \\ 
billing & condition & donate & guarantee & method & post & regard & scan & track &  \\ 
birth & confirm & drop & help & money & prefer & register & screenshot & tracking &  \\ 

\bottomrule
\end{tabular}
}%
\caption{Important lemmas utilized for template selection.}
\label{tab:lemmas}
\end{table*}

\begin{algorithm}[htbp]
	\small
	\caption{Response Pool Generation Process}
	\begin{algorithmic}[1]
		\State Preprocess data by sentence splitting, tokenization, part-of-speech tagging lemmatization.  
		\State Transform each sentence into a sequence of \texttt{verb}, \texttt{noun}, \texttt{adjective}, \texttt{adverb} lemmas by dropping punctuation and non-content words of other parts of speech.
		\State Manually review top 1k frequent \texttt{verb} and \texttt{noun} lemmas to retain a list of keywords $kw$. \Comment{We kept altogether 215 lemmas that can be found in Table \ref{tab:lemmas}, with \textasciitilde{}30 minutes of manual effort.}
		\State Template set $T = \emptyset$
		\For{sentence $s \in$ dataset} {\it [in decreasing order of frequency]}
		\For{sentence $t \in T$}
		\State $sim(s, t)=exp(\frac{\sum_{k=1}^2 ln(J_k(s_{n}, t_{n}))}{2})$
		\EndFor
		\State $sim(s, T) = argmax_{t \in T}(sim(s, t))$
		\If{$sim(s, T)<\lambda$}
		\If{$freq(s)>f_1$} 
		\State $T =T+ \{s\}$
		\EndIf
		\If{$freq(s)>f_2$ \& $s \cap kw \neq \emptyset$}
		\State $T =T+ \{s\}$
		\EndIf
		\EndIf
		\EndFor
		\State Manually remove sentences from $T$ that have grammatical errors or are inappropriate for usecase (e.g.~greetings).
		\Comment{We used $\lambda=0.4$, $f_1=350$, $f_2=15$. $J_k$ denotes Jaccard similarity of unigrams ($k=1$) and bigrams($k=2$)}
	\end{algorithmic}
	\label{tempgen_algo}
\end{algorithm}

\section{User Feedback} \label{sec:appd}

\begin{table*}[htpb]
\resizebox{\linewidth}{!}{%
\scriptsize
\begin{tabular}{ll} \hline
CHATBOT: & You can also leave a comment about how your experience went. This helps me improve.\\
USER: & Thanks so much!!! I was afraid I'd not get a refund, let alone get a return. Thanks so very much \\ \hline \hline
CHATBOT: & Again, I am sorry for the trouble that you had faced due to this circumstance and but for now do you have some clarification \\ 
& or further question regarding with my resolution? \\
USER: & No thank you \\ \hline \hline
CHATBOT: & How does that sound? \\
USER: & Sounds ok.....is there another product you would recommend that will work?
\\\hline \hline
CHATBOT: & Would there be anything else I can do to help? \\
USER: & no. Thanks for your help Mr. or Mrs. Bot \\

\hline
\end{tabular}
}%
\caption{Examples for positive user feedback.}
\end{table*}

\twocolumn
\section{Template Exploration}\label{sec:appc}
One of limitations that we observed from model online deployment is that our deterministic models always to try to rank the high volume templates. This gives less opportunity to the templates with lower frequency. For instance, for some templates with similar semantic meaning, ``\textit{No worries, let me see what I can do to help you out}" is a high frequency template, while ``\textit{No worries, I will check that for you real quick.}''  is a low frequency template. To improve the diversity of templates, we enable the deterministic ranking models with exploration capability by using Gumbel-Softmax Trick~\cite{JangETAL:17,MaddisonETAL:17}, which is originally proposed to make discrete variables to be differentiable. Here we only use the sampling functionality with temperature to control the degree of exploration. The main idea is to replace the original sigmoid score function with Gumbel-Softmax. For each inference run, we sample a template based on the computed scores.

\subsection{Dataset and Implementations}
We collected the human-in-the-loop data from deployed deterministic ranking model and exploration model. Over around 8 weeks, we collected $67,136$ samples as training set, $5000$ and $8196$ samples as validation and test set respectively, for each model. 
To make a fair comparison, we ensure the evaluated sets for each model are same. We have totally three types of datasets for evaluation: (1) $\mathcal{D}_A$: the test set is a combination of the test set of deterministic SFT data, exploration SFT dataset and general conversation SST dataset. (2)$\mathcal{D}_B$: SST test dataset; (3) $\mathcal{D}_C$: SST validation dataset.

We use the original ranking setting for experiment as described in Sec.\ref{sec:tempgen}, and set temperature=1. We want to examine: (1) how fast the exploration model can help to explore and lift those tail templates; (2) and the predictive performance of exploration model as compared to deterministic model.
\subsection{Exploration Results}
\begin{figure}
\centering
\includegraphics[width=0.5\textwidth]{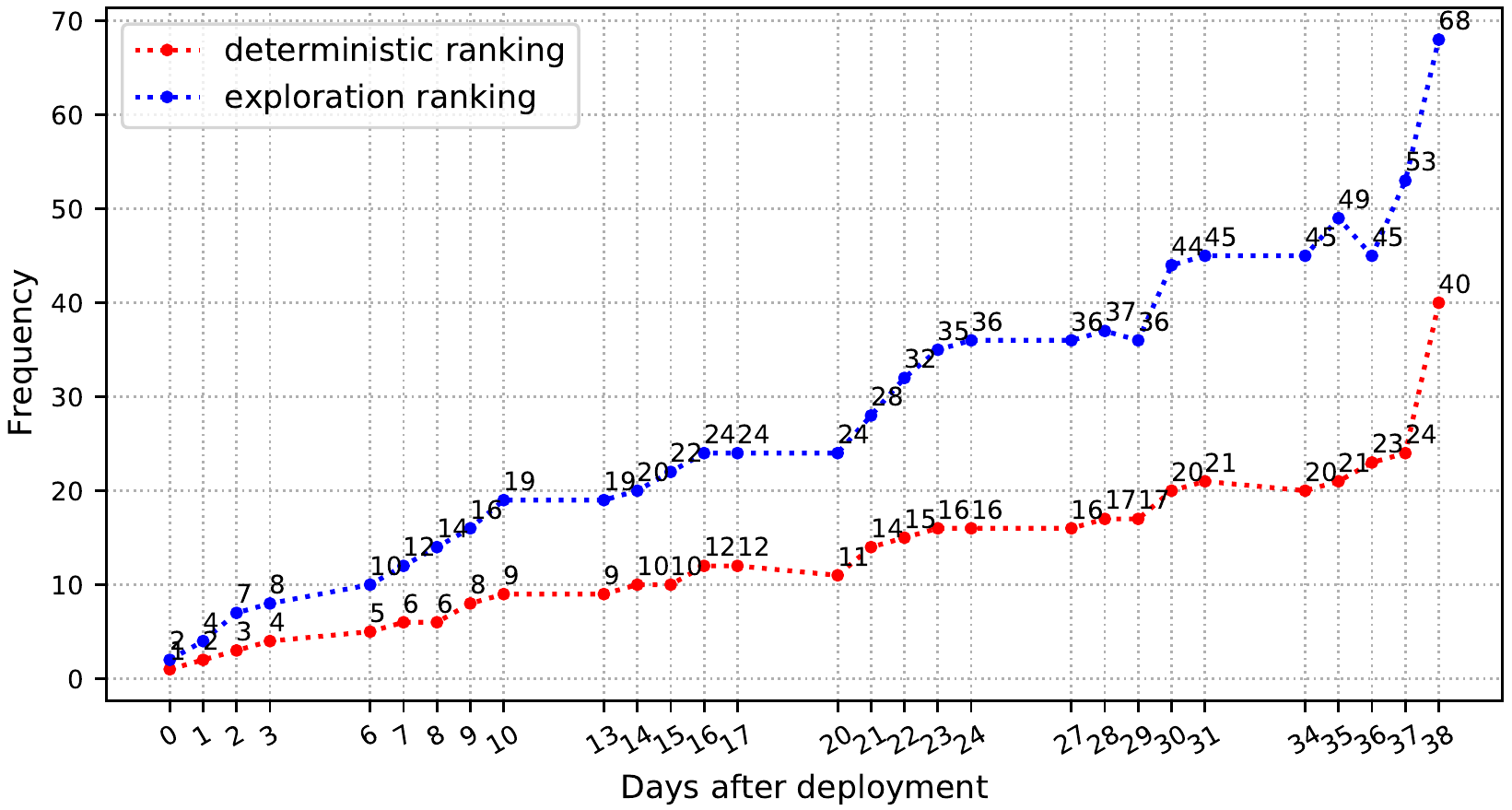}
\caption{The median of cumulative frequency of templates over time (days) for deterministic and exploration models.}
\label{fig:exploration}
\end{figure}

\begin{table}[htpb]
\centering
\small
\begin{tabular}{lcccc} \toprule
Model / Dataset & Loss & Acc & Recall@1 & MRR \\
\midrule
$\mathbf{M}_{d} (\mathcal{D}_A)$  & 0.2186 & 0.9203 & 0.9430 & 0.9707 \\
$\mathbf{M}_{e} (\mathcal{D}_A)$  & 0.2161 & 0.9197 & 0.9434 & 0.9710 \\ \midrule
$\mathbf{M}_{d} (\mathcal{D}_B)$  & 0.2284 & 0.9161 & 0.9397 & 0.9691 \\
$\mathbf{M}_{e} (\mathcal{D}_B)$ & 0.2257 & 0.9163 & 0.9397 & 0.9691 \\ \midrule
$\mathbf{M}_{d} (\mathcal{D}_C)$ & 0.1756 & 0.9322 & 0.6094 & 0.7547 \\
$\mathbf{M}_{e} (\mathcal{D}_C)$  & 0.1749 & 0.9320 & 0.6162 & 0.7584 \\
\bottomrule
\end{tabular}
\caption{The performance comparison between deterministic ranking model ($M_{d})$ and exploration ranking model ($M_{e}$) on different evaluated datasets.}
\label{tab:exploration}
\end{table}

Table \ref{tab:exploration} presents the accuracy, Recall@1 and MRR performance for each model. As we can see, exploration ranking model doesn't hurt the original predictive performance when performing exploration on templates. This is important in real-world setting, because the degraded model performance usually leads to unsatisfactory customer experience.

Figure \ref{fig:exploration} demonstrates that the exploration ranking model helps to generate a target $K$ impression for the average template ~2-3 faster than that of deterministic ranking model.

\end{document}